\documentclass{article}
\usepackage[final]{corl_2020} 
\usepackage[utf8]{inputenc}
\usepackage{xspace}

\title{Towards General and Autonomous Learning of Core Skills: A Case Study in Locomotion}

\author{
Roland Hafner \\
DeepMind \\
\texttt{rhafner@google.com} \\
\And
Tim Hertweck \\
DeepMind \\
\texttt{thertweck@google.com} \\
\And
Philipp Klöppner \\
TU Darmstadt\\
\texttt{philipp.kloeppner@gmx.de}
\And
Michael Bloesch \\
DeepMind \\
\texttt{bloesch@google.com} \\
\And
Michael Neunert \\
DeepMind \\
\texttt{neunertm@google.com} \\
\And
Markus Wulfmeier \\
DeepMind \\
\texttt{mwulfmeier@google.com} \\
\And
Saran Tunyasuvunakool \\
DeepMind \\
\texttt{stunya@google.com} \\
\And
Nicolas Heess \\
DeepMind \\
\texttt{heess@google.com} \\
\And
Martin Riedmiller \\
DeepMind \\
\texttt{riedmiller@google.com} \\
}

\usepackage{mathtools}
\usepackage{amsmath}
\usepackage{amssymb}
\usepackage{bm}
\usepackage[mathscr]{euscript}
\usepackage{natbib}
\usepackage{graphicx}
\usepackage{subfig}
\usepackage{wrapfig}
\usepackage{hyperref}
\hypersetup{
    colorlinks=true,
    linkcolor=blue,
    filecolor=magenta,      
    urlcolor=cyan,
}

\newcommand{\DAISY}{Daisy\xspace}
\newcommand{\DAISYIII}{Daisy3\xspace}
\newcommand{\DAISYIV}{Daisy4\xspace}
\newcommand{\DAISYVI}{Daisy6\xspace}
\newcommand{\DOG}{Dog\xspace}
\newcommand{\FLORENCE}{Florence\xspace}
\newcommand{\FLORI}{Flori\xspace}
\newcommand{\FLORIARMS}{FloriArms\xspace}

\newcommand{\TASKUPRIGHT}{\textit{StandUpright}\xspace}
\newcommand{\TASKLIFTFOOT}{\textit{LiftFoot}\xspace}

\newcommand{\TASKTURN}{\textit{Turn}\xspace}
\newcommand{\TASKTURNLEFT}{\textit{TurnLeft}\xspace}
\newcommand{\TASKTURNRIGHT}{\textit{TurnRight}\xspace}

\newcommand{\TASKWALK}{\textit{Walk}\xspace}
\newcommand{\TASKWALKFWD}{\textit{WalkForward}\xspace}
\newcommand{\TASKWALKBWD}{\textit{WalkBackward}\xspace}
\newcommand{\TASKWALKLEFT}{\textit{WalkLeft}\xspace}
\newcommand{\TASKWALKRIGHT}{\textit{WalkRight}\xspace}

\newcommand{\TASKREACH}{\textit{ReachTarget}\xspace}

\begin{document}

\maketitle

\begin{abstract}
Modern Reinforcement Learning (RL) algorithms promise to solve difficult motor control problems directly from raw sensory inputs.
Their attraction is due in part to the fact that they can represent a general class of methods that allow to learn a solution with a reasonably set reward and minimal prior knowledge, even in situations where it is difficult or expensive for a human expert.
For RL to truly make good on this promise, however, we need algorithms and learning setups that can work across a broad range of problems with minimal problem specific adjustments or engineering.
In this paper, we study this idea of generality in the locomotion domain.
We develop a learning framework that can learn sophisticated locomotion behavior for a wide spectrum of legged robots, such as bipeds, tripeds, quadrupeds and hexapods, including wheeled variants.
Our learning framework relies on a data-efficient, off-policy multi-task RL algorithm and a small set of reward functions that are semantically identical across robots.
To underline the general applicability of the method, we keep the hyper-parameter settings and reward definitions constant across experiments and rely exclusively on on-board sensing. For nine different types of robots, including a real-world quadruped robot, we demonstrate that the same algorithm can rapidly learn diverse and reusable locomotion skills without any platform specific adjustments or additional instrumentation of the learning setup.
(Supplementary \href{https://youtu.be/7V0-oj3b5I4}{video} available.\footnote{\href{https://youtu.be/7V0-oj3b5I4}{https://youtu.be/7V0-oj3b5I4}})
\end{abstract}

\section{Introduction}
Robots with legs and hybrid leg-wheel configurations are rapidly gaining popularity as mobile helpers with the potential to navigate challenging, human-built environments. The control of such platforms is a well studied engineering and research problem and the last two decades have seen impressive demonstrations of robot locomotion, with solutions excelling both in speed and robustness \cite{hutter2014quadrupedal, feng2015optimization, di2018dynamic, bjelonic2019keep}. Despite these successes, however, approaches used by classical control engineers and roboticists usually require detailed understanding of and specialization to the platform at hand.
Learning based approaches to control, especially Reinforcement Learning (RL) algorithms, have made much progress in the last few years \cite{tuomas18, sacx1, heess2017emergence, hwangbo2019learning}. They hold the promise of solving challenging motor control problems directly from raw sensory inputs, optimizing the perception-action pipeline end-to-end. 
In particular, they can be a general paradigm that allows us to learn a solution, even if it were difficult or expensive for a human expert, with a well-defined goal but minimal prior knowledge.
However, in order for RL to really keep this promise, we need algorithms and learning setups that can function across a wide range of problems with minimal problem-specific adjustments or design.
Although the data-efficiency and robustness of RL algorithms has much improved,
significant task-specific effort is still required for algorithm tuning, reward design and providing specific hard- and software for reward calculation.
This can make the success of learning experiments highly dependent on the availability of RL expert knowledge and limit them to carefully controlled lab settings.
Our learning framework relies on a data-efficient multi-task RL algorithm \cite{sacx1}.
With a small set of reward functions that are semantically simple and, above all, identical across robots, we show that we can learn sophisticated locomotion behavior for a wide range of robots such as bipeds, tripeds, quadrupeds and hexapods, including wheeled variants. 
We demonstrate that in our learning framework, the same RL agent, with a single setting of hyperparameters and the same set of reward functions, can learn diverse and reusable locomotion skills for 9 different types of robots.
The framework is sufficiently data efficient to enable learning directly on a real-world quadruped without any adjustments.
Although the reward semantics are identical across robots the resulting control policies vary significantly in line with the highly diverse dynamic properties of the platforms. 
Importantly, as it relies exclusively on on-board sensing it does not require any additional instrumentation of the learning setup, neither for state estimation for the controller nor for reward calculation, thus enabling learning experiments beyond a controlled lab setting.

Our results are complementary to other recent results on learning locomotion such as those of \citep{tuomas18,FacebookDaisy19,WalkInRealWorld20}, which focus primarily on data-efficiency, robustness of the resulting gaits, or the autonomy of the learning process. 
Our work also addresses these points but specifically emphasizes the generality and robustness of the learning framework.
%
Beyond locomotion, and in combination with the results of \citep{sacx1, hertweck2020simple} the results in the present paper provide another small piece of evidence that the grand vision of general, autonomous robot learning may not be entirely beyond reach. 

\section{Preliminaries}
The goal of this paper is to study the generality of learning techniques and we thus want to evaluate our learning framework on a diverse set of robot platforms.
To reduce the effort, we will mainly work with platforms that are simulated as true to the original as possible.
Unfortunately, even creating and validating a large number of independent simulation models requires a lot of work.
We therefore rely on a modular hardware system which allows to construct different robot models from a small number of hardware building blocks.
Rather than performing system identification for each robot model separately, we can then identify the properties of the hardware modules in isolation and use these well-calibrated simulation components to build a large number of realistic models with very different morphologies and dynamic properties.
Obtaining a good alignment between the learning results in the simulation and the actual hardware on a small number of models can give us some confidence that the results in the simulation are meaningful for other models as well.\par
\begin{figure*}[!ht]
\centering

\subfloat[Real Daisy.\label{fig:daisy_real}]{
\includegraphics[height=0.14\textwidth]{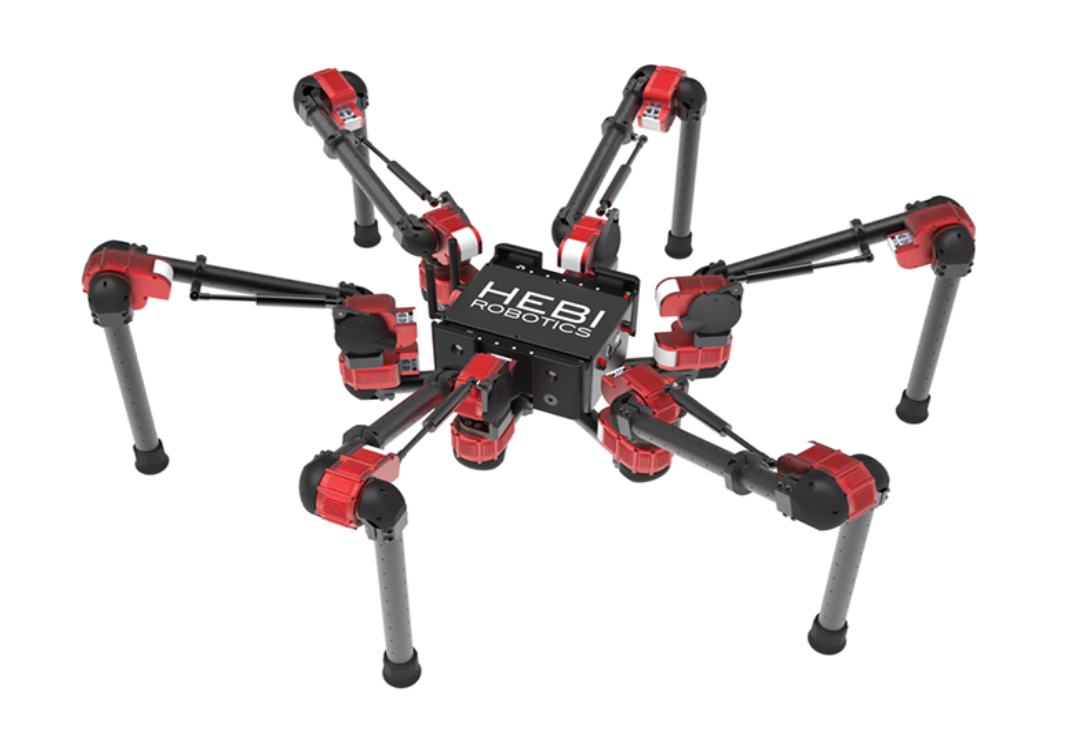}
}
\subfloat[Real Florence.\label{fig:florence_real}]{
\includegraphics[height=0.14\textwidth]{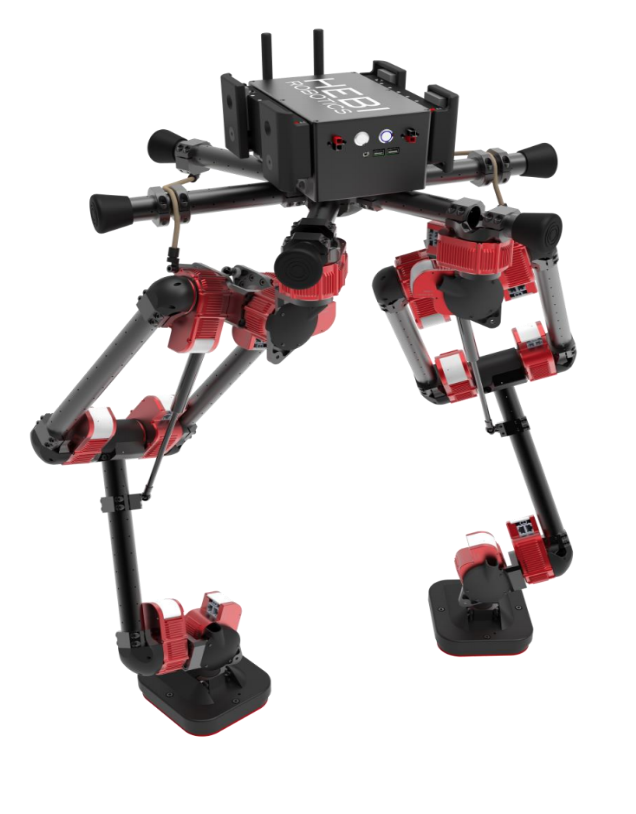}
}
\subfloat[\DAISYIII \label{fig:daisy3_sim}]{{
\includegraphics[height=0.14\textwidth]{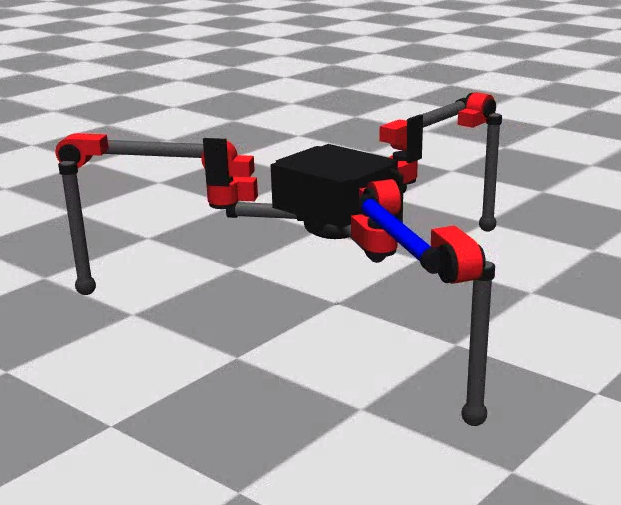}
}}
\subfloat[\DAISYIV \label{fig:daisy4_sim}]{
\includegraphics[height=0.14\textwidth]{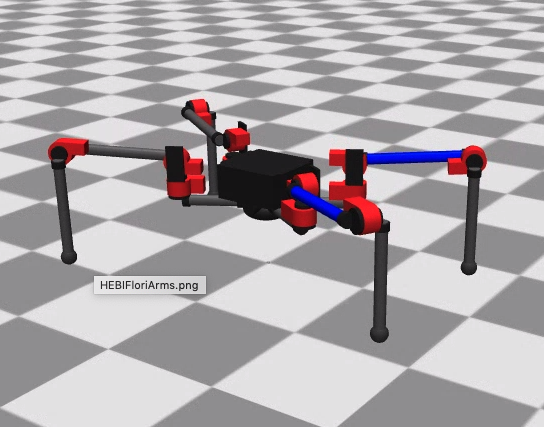}
}
\subfloat[\DAISYVI \label{fig:daisy6_sim}]{
\includegraphics[height=0.14\textwidth]{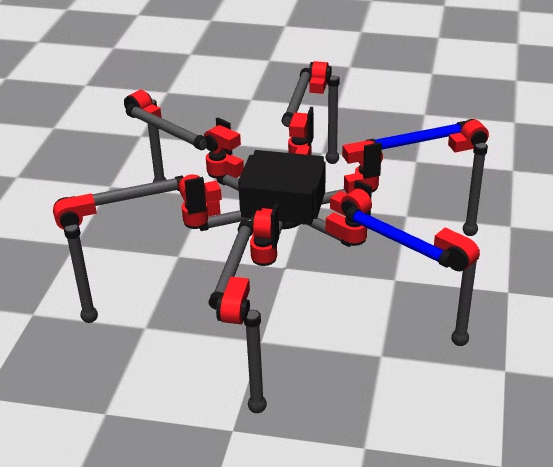}
} \\
\subfloat[\DOG \label{fig:dog_sim}]{
\includegraphics[height=0.14\textwidth]{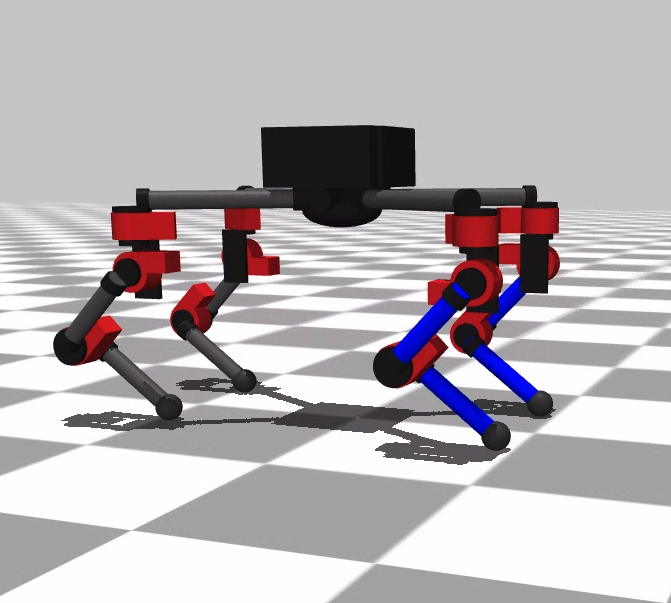}
}
\subfloat[\FLORENCE \label{fig:florence_sim}]{
\includegraphics[height=0.14\textwidth]{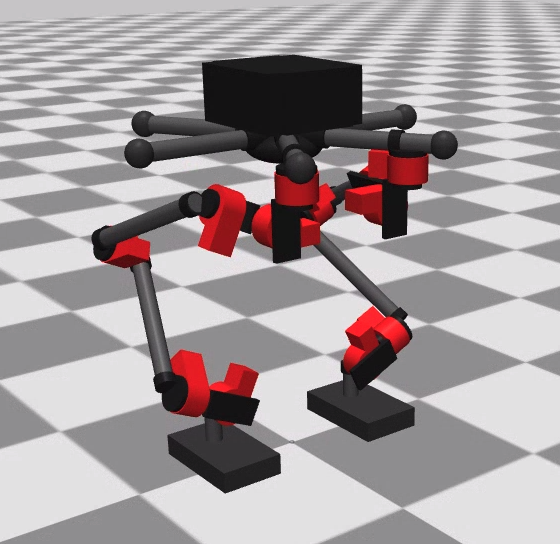}
}
\subfloat[\FLORI \label{fig:flori_sim}]{
\includegraphics[height=0.14\textwidth]{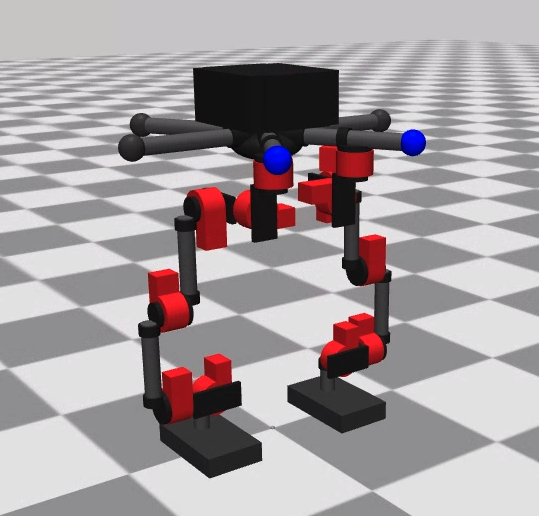}
}
\subfloat[\FLORIARMS \label{fig:flori_arms_sim}]{
\includegraphics[height=0.14\textwidth]{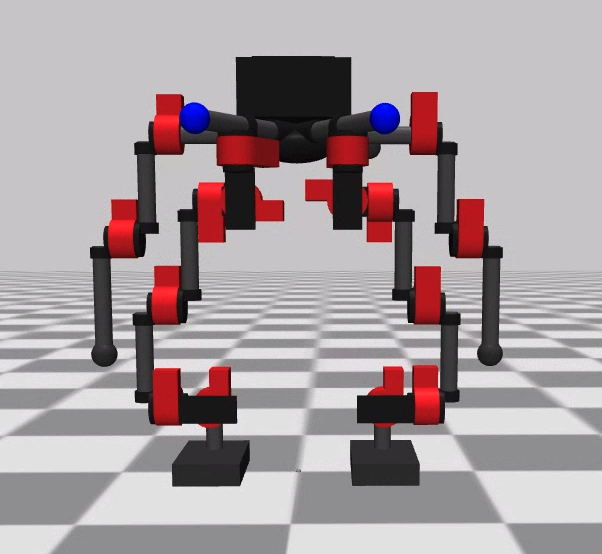}
}
\subfloat[Real \DAISYIV \label{fig:daisy4_real}]{
\includegraphics[height=0.14\textwidth]{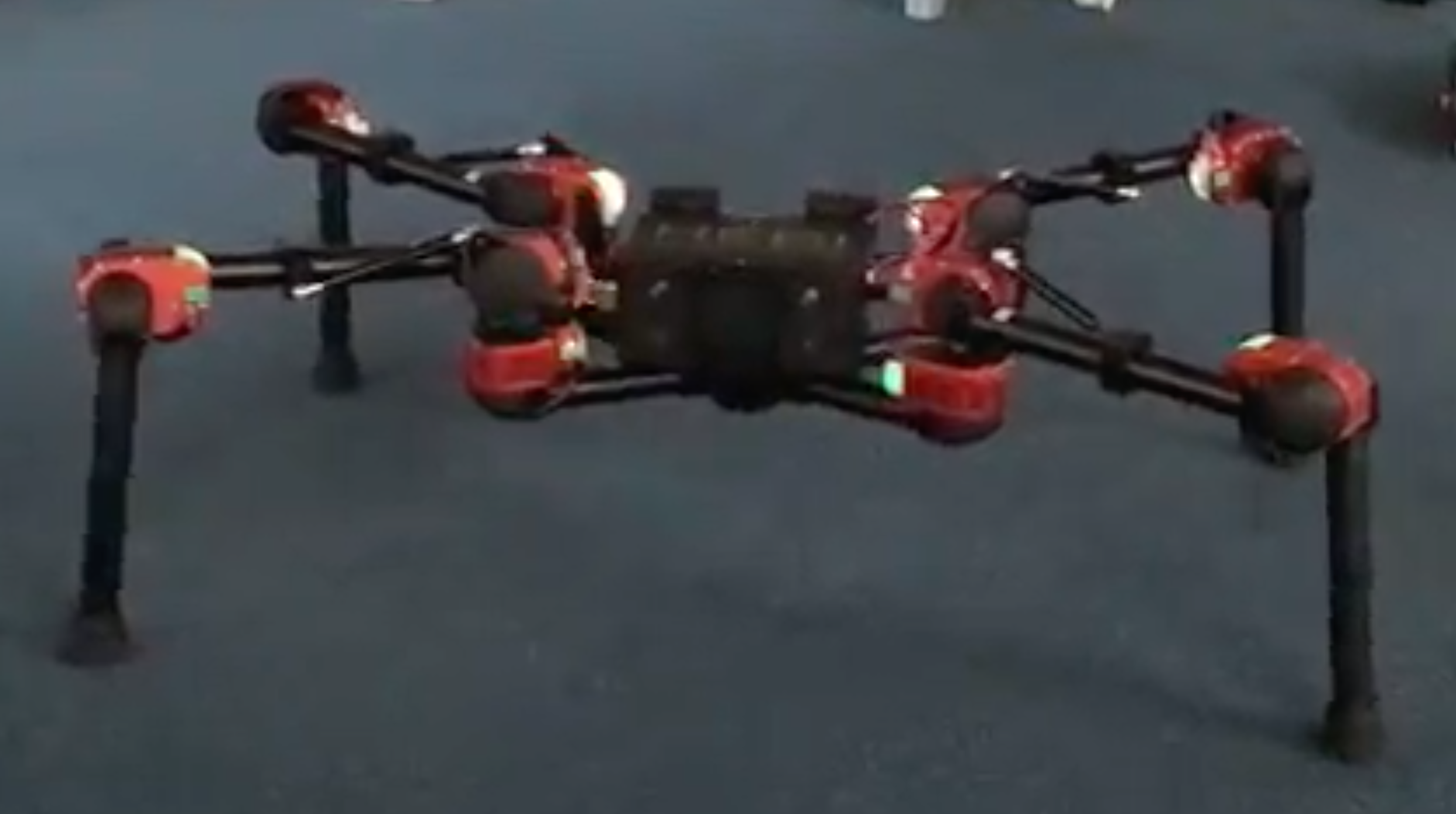}
}
\caption{Real and simulated creatures built using the HEBI system. \label{fig:creatures_sim}}
\end{figure*}

HEBI Robotics (www.hebirobotics.com) is a provider of a modular hardware system for robotics.
The system is built around series elastic actuator modules that are available with different nominal speed and torque ratings.
The series elastic elements allow for accurate torque control and protect the motor from strong impacts. The modules have a rich set of sensors built in: encoders for the motor and the output shaft, temperature sensors, a 3 axis accelerometer, as well as a 3 axis gyroscope (including on board orientation estimation).
A low level controller, consisting of integrated control and power electronics, implements different control modes and safety mechanisms, and processes sensor information.
In combination with accessories such as brackets and tubes, the modules allow building a wide variety of different robots, including (but not limited to) legged robots.
An attractive feature of the system is that we have access to all relevant state variables used for low-level control (actuator position, velocity, deflection, deflection velocity, torque, motor temperature, etc.).
In Fig \ref{fig:florence_real} and \ref{fig:daisy_real} two existing walker topologies are shown.
As \FLORENCE is currently only available as a prototype we use the \DAISY kit for evaluation of our learning framework in real-world experiments.\par
\label{sec:sim}%
For our investigations, we have developed a MuJoCo \cite{todorov2012mujoco} based simulation of these building blocks.
In our simulation we attempt to faithfully reflect the modularity as well as the kinematic and dynamic properties of the system.
The latter include not only the properties of the motor, gearbox and serial elastic element, but also the firmware safety features, temperature models and overheating effects.
We implemented 7 basic robot models with different morphologies and dynamic properties.
An overview can be seen in Fig \ref{fig:daisy3_sim} to \ref{fig:flori_arms_sim}).

Fig \ref{fig:daisy6_sim} shows the Daisy hexapod (\DAISYVI) in the configuration of the original kit.
It has two degrees of freedom in each shoulder and one additional in each elbow, resulting in 18 active degrees of freedom.
By removing two legs, we get a more challenging to control quadruped robot (\DAISYIV, see Fig \ref{fig:daisy4_sim}), with 12 active degrees of freedom.
It is notable that for a human engineer, these two topologies already differ in important dynamical aspects: e.g. only the hexapod allows for a simple bipartite statically stable gait \cite{saranli2001rhex}.
To have non-optimal, but still somehow feasible, kinematics, we can remove another leg to get a three legged version of the Daisy (\DAISYIII, see Fig \ref{fig:daisy3_sim}), with nine active degrees of freedom (which is very difficult to control with standard methods since it will almost inevitably have to rely on friction dynamics).

We also modeled a mammalian leg configuration (see Fig \ref{fig:dog_sim}), by changing the orientation of the shoulders and adapting slightly the lengths of the upper and lower legs.
This assembly has the same number of active degrees of freedom as \DAISYIV, but strongly differs in forward kinematics and joint torque loads (it is usually less strenuous for the shoulder joints).

Another important class of robots for locomotion are bipedal robots.
Fig \ref{fig:florence_sim} shows a simulation model of HEBI \FLORENCE.
\FLORENCE has three active degrees of freedom in each hip, one degree of freedom in the knee and two degrees of freedom in each ankle.
It further has long upper and lower leg segments and starts with a backward flexed knee to prevent operation close to kinematic singularities and to make balancing easier.
While the latter are design choices that were made with engineered control solutions in mind, we also designed a more human like version of the Florence - that we call \FLORI (see Fig \ref{fig:flori_sim}) - with the same number of active degrees of freedom but different leg configurations.
To have a rudimentary example for additional limbs, we also added a \FLORI version with two arms adding two additional degrees of freedom for each of them (\FLORIARMS, see Fig \ref{fig:flori_arms_sim}).

\section{A General Framework for Learning Core Locomotion Skills}
\label{sec:core_controller}

Our long-term goal is the development of a general and autonomous learning framework. `General' means that the same framework with minimal modifications can be applied across a broad range of different platforms. `Autonomous' means, that our system is able to learn with minimal external infrastructure and assistance. In the present work, we focus on general proprioceptive rewards, relying only on on-board sensing, thereby reducing reliance on external sensors, and other elaborate lab settings.

Many contemporary applications of RL require careful, task-specific engineering of the rewards together with expensive additional hardware such as motion capture systems, to enable reward computation. This can make learning experiments expensive and often restricts them to specialized laboratories.
Reducing this dependency both for acting and learning will increase the applicability of mobile robots, and, importantly, dramatically simplify the setup of learning experiments, enabling learning and adaptation to proceed after deployment.

Rather than relying on pre-processed position or velocity information from a separate state estimation system, we train agents to act directly from raw sensor values. We further demonstrate how a set of primitive rewards for locomotion can be derived directly from the same on-board sensors also used for acting, and how these rewards can be used to learn diverse and robust locomotion skills.

\subsection{Reward Computation for General Locomotion Topologies}
\label{subsec:rewards_general}
Our learning framework relies on a diverse set of basic rewards. The combination of these rewards enables learning a diverse set of locomotion skills which can subsequently help to learn more complex behaviors. The rewards are defined such that they can be computed from limited on-board sensing comprising IMUs and joint encoders but require no contact sensing. To this end we draw on heuristics to obtain rough estimates of helpful quantities such as egocentric velocity. The underlying assumptions of these estimates do not hold at all times, but the agent has access to the full sensory stream and is able to learn robust locomotion skills despite the potentially limited consistency of the rewards. 
Assuming that the lowest foot is in contact with the ground and not slipping, we estimate the linear velocity of the robot torso and the feet in a coordinate system that is simultaneously aligned with the robots forward direction and gravity (motivated by \cite{bloesch2017technical}, details see Appendix).
Together with the IMU gyroscope measurements in the torso, we can now define various rewards for locomotion.

An important skill in locomotion is to learn to stand upright.
We define the \textbf{\TASKUPRIGHT} reward
by keeping the robot torso leveled (reducing the roll and pitch angles) while keeping the torso linear velocity and the torso angular rotation rate small.
If we add a component for rewarding height differences of a certain foot $i$ w.r.t the lowest foot, we can define a reward function for standing upright and lifting a certain foot: \textbf{$\TASKLIFTFOOT_i$}.
For doing actual locomotion, we can modify the stand upright reward by rewarding rotational velocities around the torso z axis to get a \textbf{\TASKTURN} reward
and reward translational velocities of the torso and the feet to get a \textbf{\TASKWALK} reward.
While we could have rewards for different velocities, we picked rewards to maximize discrete instances of these rewards for this paper. In consequence we define six distinct locomotion skill rewards: \textbf{\TASKTURNLEFT}, \textbf{\TASKTURNRIGHT}, \textbf{\TASKWALKFWD}, \textbf{\TASKWALKBWD}, \textbf{\TASKWALKLEFT} and \textbf{\TASKWALKRIGHT}, as well as the \textbf{$\TASKLIFTFOOT$} for all feet for each creature (for details, see Appendix).

\subsection{Action and Observation Space}
The actuation modules offer multiple control modes, including position control, velocity control, torque control and PWM direct control mode.
In principle, our learning methods should be able to cope with all of these modes and will learn to make use of them.
While each of the modes has its own pros and cons, we picked the position control mode using a low-gain P-controller.
The main advantage of the position control mode is that we can enforce certain limits of the joint angles during the execution of our agent while it can still regulate forces indirectly by choosing appropriate position set-points.
As an additional safety mechanism, we use a sliding window filter with a width of $\nu$ steps for the set-point that is sent to the actuation modules.
In consequence, we have an action space where each of the used modules adds one dimension of continuous actions that is bounded by the allowed position set-point for that individual joint.

\label{sec:obs_space}%
For a robot that is built from multiple actuator modules, the observation space consists of observations associated with the individual modules, as well as observations from the torso.
For a default filter window of width $\nu=5$, this adds up to a 11 dimensional observation for each of the actuation modules, containing the position and velocity of the joint and elastic element, temperatures and filter state.
For the torso observations, we stack measurements of $h$ consecutive time frames to allow the agent to have richer information about the state.
As we only use robo-centric measurements, we provide the roll and pitch estimate together with the feet reference points and the measurements of the gyro.
Consequently, the range of dimensionality of the action and observation spaces we investigate here ranges from 9 action dimensions with 127 observation dimensions for \DAISYIII up to 18 action dimensions for \DAISYVI and 282 observation dimensions for \FLORIARMS (details, see Appendix).

\subsection{Multi-Task Training of a Locomotion Module}
\label{multi_task_learning}

In general, we aim for a capable locomotion module, that can not only solve one task, but is able to perform multiple tasks.
This makes the motion module not only more versatile, we also expect synergies across tasks that will improve data-efficiency in this multi-task learning setting.
To this end, we apply the Scheduled Auxiliary Control (SAC-X) \citep{sacx1} framework to the domain of locomotion.
The core idea of SAC-X is that we can learn multiple tasks in parallel, switching between different tasks during each episode, and sharing data across tasks for learning. 
This framework has three potential advantages: (1) switching between tasks forces the agent to visit different parts of the state space and can thus improve exploration (and in consequence data-efficiency); (2) switching between tasks can also improve robustness of policies since behaviors are initiated in a more diverse set of states; (3) sharing data across tasks via off-policy learning can further improve data-efficiency. We expect the resulting controller module to provide a sound basis of finely tuned movement skills that eventually also allow to achieve more high-level goals.

\section{Experiments}
To investigate the framework outlined in the previous section we conduct a case study that focuses on a set of basic locomotion skills
\TASKUPRIGHT, \TASKLIFTFOOT, \TASKTURNLEFT, \TASKTURNRIGHT, \TASKWALKLEFT, \TASKWALKRIGHT, \TASKWALKFWD, \TASKWALKBWD (see \ref{subsec:rewards_general}; note that additional rewards could be easily defined following the same approach).
We use the off-policy RL algorithm used in \citep{sacx1}, using the very same hyper-parameters that were also used in other domains like manipulation.
In each episode the robot starts with all actuators in the default position, feet touching the ground (see Fig \ref{fig:daisy6_sim} to \ref{fig:florence_sim}).
We run each episode for 800 steps with a control time step duration of 25 ms, which yields episodes of 20 seconds length.
We are interested in applying our approach directly on a real robot platform. Our main interest therefore is data-efficiency
which we measure by counting the episodes that were required to learn the behaviour(s) (details, see Appendix).
This gives us a good estimate of whether learning the tasks has reached a level of efficiency such that it could be trained in the real world.

\subsection{Individual Skills}
\label{sec:single_task_experiments}

We first investigate the plausibility of our reward definition in a singe-task setting.
As Table \ref{table:results_single_task} shows, we can learn the individual locomotion skills on all platforms in a reasonable number of interaction episodes.
For instance, starting from a random policy we can successfully learn behaviours like \TASKUPRIGHT for creatures like \DAISYVI, \DAISYIV and \DAISYIII in less than 20 interaction episodes.
This is equivalent to less than 7 minutes of interaction between the agent and the robot.
Furthermore, our results for the bipedal robots \FLORENCE, \FLORI and \FLORIARMS show that the very same reward definition can have a very different complexity depending on the configuration we apply it to.
Since the static stability of these creatures is strongly impeded by the reduced support polygons, the agent needs to be much more careful when moving its center of mass. Still it can learn the task in less than 4h of interaction time (about 700 episodes) for all robots.

This is even more evident for \TASKLIFTFOOT: depending on the structure of the robot platform the same reward definition results in tasks of varying difficulties and leads to very different solution strategies:
While we can learn to lift a certain foot for \DAISYVI, \DAISYIV and \DOG in less than 20 minutes of interaction time (about 50 episodes), the task is considerably harder for the bipedal robots. Nevertheless, the very same agent and reward learns a balancing policy on one leg in about 5.6h of interaction time (about 1000 episodes).
\begin{table*}[h!]
\centering
 \begin{tabular}{|| c || c || c || c || c || c | c ||}
  \hline
  \multicolumn{1}{||c||}{} & \multicolumn{1}{c||}{\TASKUPRIGHT} & \multicolumn{1}{c||}{$\TASKLIFTFOOT_1$} & \multicolumn{1}{c||}{\TASKTURNLEFT} & \multicolumn{1}{c||}{\TASKWALKLEFT} & \multicolumn{2}{c||}{\TASKWALKFWD}\\
  \hline
   & Episodes & Episodes & Episodes & Episodes & Episodes & Velocity \\
  \hline \hline
   \DAISYVI   & $<$20 &  60    & 120   &  180  &  170  &  0.59 m/s\\
   \DAISYIV   & $<$20 &  50    & 90    &  150  &  160  &  0.62 m/s\\
   \DOG       & 30    &  50    & 210   &  520  &  310  &  0.70 m/s\\
   \DAISYIII  & $<$20 &  N/A   & 420   &  250  &  240  &  0.01 m/s\\
   \FLORENCE  &  650  &  1000  & 1100  &  1340 &  1200 &  1.20 m/s\\
   \FLORI     &  710  &  1000  & 1220  &  1220 &  1120 &  1.45 m/s\\
   \FLORIARMS &  700  &  320   & 1400  &  1210 &  1100 &  1.44 m/s\\
   \hline\hline
 \end{tabular}
 \caption{\label{table:results_single_task}Results for learning basic locomotion tasks on different walker platforms in a single-task setting. Shown are the interaction episodes required to learn the task and the final velocity reached for the \TASKWALK tasks.}
\end{table*}
These results highlight that the same simple reward definition can give rise to very different behaviors. 
We get comparable results for \TASKTURNLEFT and \TASKTURNRIGHT, as all of our creatures are symmetric.
For \DAISYIV and \DAISYVI these tasks are considerably more difficult than \TASKUPRIGHT and \TASKLIFTFOOT and the amount of interaction data that is required to learn the skills roughly doubles.
For \TASKWALK we can learn a reasonable fast solution for \DAISYVI and \DAISYIV in about an hour of interaction time.
The resulting gait looks highly symmetric
even though we do not directly encourage this in the reward.
Interestingly, the agent also finds a very good gait for the bipeds \FLORENCE, \FLORI and \FLORIARMS in about 7.5h of interaction time (about 1200 episodes). This walking gait looks not only very symmetric but also very dynamic.
The learned walking gait for \TASKWALKLEFT, \TASKWALKRIGHT, \TASKWALKFWD and \TASKWALKBWD take a comparable amount of interaction episodes to learn, but as can be seen in Table \ref{table:results_single_task} vary widely in the achievable speed.

It is worth noting that we apply exactly the same reward function, agent and hyperparameters to all robot platforms. The characteristics of the resulting behaviors, however, vary widely and are naturally adapted to the morphology and dynamic properties of each platform, e.g. \FLORIARMS learns to use its arms for additional support while lifting a leg and to swing it's arms in a very natural way to keep balance while walking\footnote{e.g. see supplementary video \href{https://youtu.be/7V0-oj3b5I4}{https://youtu.be/7V0-oj3b5I4}}.

\subsection{Learning a Versatile Motor Module}
\label{sec:sac_experiments_without_main}
To obtain a versatile motor module we would like to be able to learn a large number of locomotion skills in parallel.
Although learning many individual skills separately is feasible, it is not the most data-efficient way to achieve this. Also, when learning skills separately we are not guaranteed to be able to transition between skills.
We therefore switch to the multi-task regime outlined in section \ref{multi_task_learning} in which we switch between and share data across tasks \citep{sacx1}.
We keep the basic learning algorithm, parameters and the general learning setting from the previous sections.
We consider three basic task definitions: \TASKWALKFWD, \TASKWALKBWD and \TASKUPRIGHT.
In every episode we execute two sequences of 10 seconds length each, giving a total episode length of 20 seconds as before.
In each sequence we randomly execute one of the three tasks to collect data (this corresponds to the SAC-U version of the algorithm described in \citep{sacx1}).

For the quadrupeds and hexapod, we see a small increase in data-efficiency compared to the single-task experiments. For example, we need 360 episodes in total for \DAISYVI when learning each task in a separate experiment, while we can learn all skills together in only 300 episodes in the multi-task setting (results are comparable for \DAISYIV and \DOG).
For the bipeds, the differences are much bigger. For example we would need 3050 episodes for  \FLORENCE to learn all skills separately, while we can learn them in 1590 episodes in the multi-task setting.
While this saves only roughly 20 minutes of interaction time for the quadrupeds and hexapod, the savings amount to over 8h of interaction time for the bipeds.
Importantly, in the multi-task setting the agent also learns to transition between \TASKWALKFWD, \TASKWALKBWD and \TASKUPRIGHT without falling, which is a very challenging task for itself for many control approaches.

\subsection{Learning Higher Level Behaviours: Reaching a Target}
\label{sec:sac_main_task}
\begin{wraptable}{r}{60mm}
\centering
 \begin{tabular}{||c c c||} 
 \hline
 Creature & Baseline & SAC-Q\\ [0.5ex] 
 \hline\hline
 \DAISYVI   & $>$20k &  840  \\
 \DAISYIV   & $>$20k &  800  \\
 \FLORENCE & $>$20k &  2000 \\[1ex]
 \hline
 \end{tabular}%
 \caption{Interaction episodes for task \TASKREACH.}%
 \label{table:results_multitask_reach}%
\end{wraptable}%
In the previous section we have demonstrated that the multi-task regime allows us to learn multiple individual skills more efficiently and robustly than when learning separately. Many more complex tasks, however, cannot reasonably be learned in a single-task setting at all.
We now show that the training regime from the previous section enables learning both locomotion skills as well as more complex tasks that build on these skills, including sparse reward tasks that would be hard to learn otherwise. 
To this end we add a virtual target to the environment that is randomly spawned in a certain range around the robot.
We add a sparse \TASKREACH task to the set of training tasks for our locomotion module. The reward is zero when the target is more than 50 cm away from the robot and one when the distance of the target and the robot torso is zero.
In each episode the target is spawned at a distance from 1 to 3 meters around the robot.

As baseline we attempt to learn the task with only the \TASKREACH reward and the default settings of our agent.
For all creatures this baseline fails to solve the task in the first 20k episodes.
As a comparison, we use our motion module in the multi-task setting together with 3 auxiliaries (\TASKWALKFWD, \TASKWALKBWD, \TASKUPRIGHT).
As we can see in Table \ref{table:results_multitask_reach}, we can learn all skills plus the main task \TASKREACH, in a reasonable time of about 5h (roughly 800 episodes) for the quadrupeds and hexapods and about 12h of interaction time (roughly 2000 episodes) for the bipeds. In this experiment, we assumed that we train all tasks from scratch, while in practice it would also be possible to pre-train a set of skills and learn only the main task, which would make the motion module even more powerful.

\subsection{Robustness of Proprioceptive Reward Definitions}
\label{sec:walking_blind}
As discussed in section \ref{subsec:rewards_general} the reward calculation is based on simplifying assumptions which may not always hold true. 
While these may appear restrictive, we do not require them to hold at every point in time in order to promote the emergence of sensible locomotion behaviors.
To demonstrate that we can deal with violations of the assumptions and to underline the robustness of our approach, we conduct experiments in an expanded set of tasks.
\begin{wrapfigure}{r}{0.52\textwidth}
\centering
\subfloat[Rough terrain.\label{fig:daisy6_tiled}]{
\includegraphics[height=0.12\textwidth]{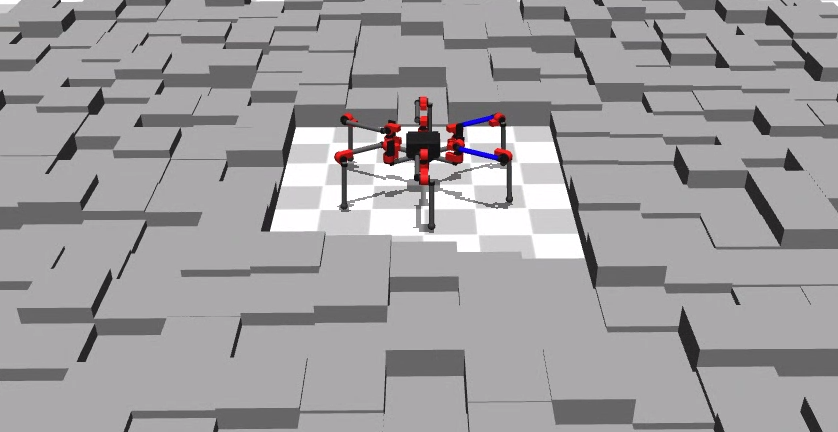}
}
\subfloat[Passive skates.\label{fig:florence_skates}]{
\includegraphics[height=0.12\textwidth]{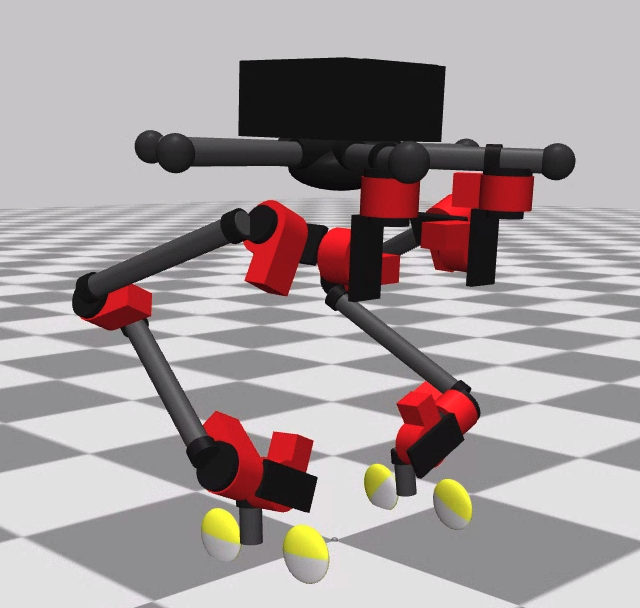}\includegraphics[height=0.12\textwidth]{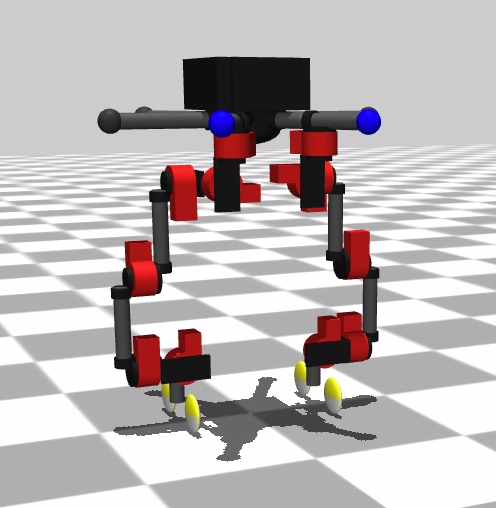}
}
\caption{Platforms in the robustness experiments.\label{fig:creatures_robust}}
\end{wrapfigure}
In a first experiment, we let our creatures run over uneven, tiled terrain and can observe that learning for height differences of a few centimeters still works successfully.
Moreover we see that platforms with more legs can overcome rougher terrain using the same rewards (e.g. see Fig \ref{fig:daisy6_tiled}).
In a different experiment, we extend the 7 creatures by 2 more and attach passive wheels to the feet of the bipeds \FLORENCE and \FLORI. Running the same experiments results in a completely different locomotion pattern: dynamic skating (see Fig \ref{fig:florence_skates}, for more details, see Appendix).

\section{Real-World Experiments}
To verify the results obtained in simulation we conduct learning  experiments `from scratch' on an actual HEBI robot.
Instead of the original HEBI Daisy (\DAISYVI, see Fig \ref{fig:daisy_real}), we decided to run the real world experiments on the more challenging quadruped Daisy4 that is shown in Fig \ref{fig:daisy4_real}.
We use the same settings as in simulation (agent, rewards, methods, hyperparameters, etc.).
From a control perspective going to a real robot means that the agent now has to deal with additional time delays and noise that makes the control problem more difficult.
When we run the single-task experiments of section \ref{sec:single_task_experiments}, we initialize the robot in each episode to its default pose by a hand designed initialization procedure.
Afterwards we can start the episode in the same way as we do in simulation.
While the reset of the robot after an episode is not a problem in simulation, we allow more time in between episodes to manually turn the robot around when it used up the available space.

To learn \TASKWALKFWD in the real robot experiment, we need approximately 130 episodes, which is even a bit less compared to the simulation experiments (160 episodes). While this corresponds to approximately 40 minutes of pure interaction time, the full experiment (including resets) runs for about 2h.
The resulting walking gait is highly symmetric and achieves a speed of approximately 0.3 m/s.
We further conduct a multi-task experiment in the real world with 6 different tasks: $\TASKLIFTFOOT_1$, $\TASKLIFTFOOT_2$, $\TASKLIFTFOOT_3$, $\TASKLIFTFOOT_4$, \TASKWALKFWD and \TASKWALKBWD.
We use the same setting as in section \ref{sec:sac_experiments_without_main}, but increase the sequence length to 20 seconds.
Starting from a random initialisation, the agent is able to learn all the tasks requiring only 225 episodes of robot interaction.
This corresponds to approximately 3h of pure interaction time, while the overall experiment (including resets) runs for about 5h. 
This demonstrates that we can learn robust skills that allow for smooth transitions not just in simulation but also on an actual robot in reasonable time from scratch. It further demonstrates that another core feature of our simulation results holds true on the real robot: specifically, the multi-task setup continues to provide us with increased data efficiency, as we would have to run 460 episodes (10h) to learn all the skills in a single-task setting. Without multi-task training we would have had to wait for additional 5h and would not have learned to transition between skills.

\section{Related Work}
Legged locomotion has seen significant progress in the last couple of decades with increasingly performing hardware and control approaches \cite{hutter2014quadrupedal, semini2011design, bledt2018cheetah}. 
Optimal control approaches, partially implemented as Model Predictive Control (MPC), have gained traction, especially combined with a centroidal dynamics approximation which allows separating the problem into a high-level base motion controller and a low-level contact force controller \cite{winkler2018gait, kuindersma2014efficiently, villarreal2019mpc}. But also whole-body approaches have been successfully investigated by various research groups \cite{koenemann2015whole, neunert2018whole}.
These approaches can reach an astonishing level of dynamics and agility \cite{BostonDynamicsAgility19}.

On the other hand side in the last few years there has been a growing interest in learning locomotion both in simulation and for real robots. In simulation, especially for simple robot models, basic locomotion behavior can often be achieved with simple reward functions \citep[e.g.][]{heess2016learning}. More sophisticated and diverse skills can be obtained
through curricula and diverse training conditions (such as different terrains) \cite{heess2017emergence}. 
However, in general, such skills require carefully chosen shaping or penalty terms \cite{hwangbo2019learning}
or constrained optimization \cite{bohez2019value} that in turn are time-consuming, may need an iterative process \cite{xie2020learning} and are specific for a certain platform. 
However, in general, such skills often lack the naturalness, efficiency, smoothness and other properties that would be essential for deployment on actual robotics hardware. 
This can be mitigated through carefully chosen shaping or penalty terms \cite{hwangbo2019learning}, or constrained optimization \cite{bohez2019value}. But designing regularization strategies that shape behaviors in particular ways can be a time-consuming endeavor and may require an iterative process \cite{xie2020learning}.

The results in this paper are also complementary to several recent demonstrations of successful sim-to-real transfer of control policies for legged robots \cite{hwangbo2019learning,xie2020learning,nachum2019multi}. Although training in simulation offers additional flexibility, successful transfer usually requires detailed knowledge of dynamic properties of the robot of interest to build sufficiently accurate simulation models or additional adaptation of the learned control strategies on the actual hardware \cite{yu2019sim,sun2018adaptive,peng2020learning}.
In some cases demonstrations, e.g.\ from motion capture data \citep[e.g.][]{peng2019mcp,merel2018neural} or other reference motions \cite{xie2020learning,yu2019sim} can be used to directly constrain learned behavior. Yet, such data is not always easily available or may not easily transfer to a particular robot body. Furthermore, composing reference behaviors in a flexible, goal-directed manner can be challenging \citep[e.g.][]{merel2018neural,peng2019mcp}. 
Our work uses a multi-task learning scheme taken from \cite{sacx1,hertweck2020simple,wulfmeier2019compositional} that employs several simple reward functions with minimal additional shaping terms to obtain well regularized and robust behavior across a number of different bodies.

In some cases, locomotion skills learned in simulation can be transferred to corresponding robotic hardware. This usually requires careful system identification and well matched simulation models \cite{hwangbo2019learning,peng2020learning,xie2020learning,nachum2019multi}. Transfer can be further improved with additional adaptation of the learned control strategies on the actual hardware \cite{yu2019sim,sun2018adaptive,peng2020learning}. 
Accurate simulation models can, however, be expensive to develop, and some phenomena encountered in the real world (such as sophisticated terrain properties) may be hard to simulate. 

Recent improvements in the efficiency of learning algorithms has made it possible to learn locomotion skills directly on the robot. This has been pursued both with model-based \cite{brain_model_handful}, and with model-free approaches \cite{tuomas18,peng2020learning,FacebookDaisy19} for quadrupeds \cite{tuomas18,peng2020learning} and the HEBI Daisy robot which we are considering here \cite{FacebookDaisy19}. Similar to our work, \cite{peng2020learning,FacebookDaisy19} learn multiple skills that can later be chained to achieve goal directed behaviors. Learning on the hardware requires answering practical questions related to safety, reset, and state-estimation e.g. to compute rewards. For the latter, prior work usually relies on external motion capture systems which can require significant effort to set up. We show that sophisticated skills can be learned from simple rewards computed from on-board sensors only, thus significantly reducing the complexity of the training setup. Furthermore, whereas prior work usually targets a single robot platform, we investigate whether the same setup can be used across a number of different robots.

Our use of multiple simple rewards derived from on-board sensing is closely related to the work of \cite{hertweck2020simple} who use a similar scheme to solve difficult tasks with a robotic arm.
It also bears similarity to a number of papers who employ learned reward functions, for instance based on an empowerment objective, to discover reusable skills \cite{gregor2016variational,hausman2018learning,eysenbach2018diversity,sharma2019dynamics,gregor2016variational}, including for legged robots \cite{sharma2020emergent}. Our reward functions are hand-crafted, but nevertheless simple and transferable across body morphologies. 


\section{Conclusion}
We have investigated a framework for learning of core locomotion skills for general walker topologies and applied it to a diverse set of robots with very different morphologies and dynamic properties. We have demonstrated that the same set of reward functions and the same learning framework (identical algorithm and hyperparameter settings) can successfully learn a diverse set of robust locomotion skills for all platforms and we can reuse these skills to learn more complex tasks.
Even though the rewards are the same for all robots, the resulting skills are naturally adapted to the characteristics of each platform. 
Our framework is sufficiently data-efficient to learn all tasks in a couple of hours of interaction time, and we have verified some of our results in simulation with matching experiments on real hardware.
Our framework and reward definitions further minimize the need for external state estimation and instrumentation of the learning setup by relying only on on-board sensing. This has already made it possible to conduct experiments for some of our robots essentially in the wild, although further work will be necessary for more complicated robots such as the biped Florence, e.g.\ to ensure their safety during learning. 

We believe that learning frameworks that are general enough to work across a wide range of platforms with minimal adjustments and that enable more autonomous learning will be an important step to fully reap the benefits of self-learning systems in robotics (for work similar in spirit in the manipulation domain, see \citep{hertweck2020simple}).

\section*{Acknowledgements}
We thank Mr Florian Enner, HEBI Robotics, for the excellent technical support and expert consultation on \DAISY and \FLORENCE.


\setlength{\bibsep}{4.5pt}
\bibliography{references}

\appendix

\newcommand{\RLD}{Reinforcement Learning (RL) }
\newcommand{\MDPD}{Markov Decision Process (MDP) }
\newcommand{\RL}{Reinforcement Learning }
\newcommand{\rl}{RL }

\newcommand{\SACD}{Scheduled Auxiliary Control (SAC) }
\newcommand{\SAC}{Scheduled Auxiliary Control }
\newcommand{\sac}{SAC }

\newcommand{\bE}{\mathbb{E}}
\newcommand{\bx}{\mathbf{x}}
\newcommand{\bs}{\mathbf{s}}
\newcommand{\ba}{\mathbf{a}}
\newcommand{\btheta}{{\bm{\theta}}}
\newcommand{\cM}{{\mathcal{M}}}
\newcommand{\sA}{\mathscr{A}}
\newcommand{\sT}{\mathscr{T}}
\newcommand{\cA}{{\mathcal{A}}}
\newcommand{\cT}{{\mathcal{T}}}
\newcommand{\cB}{{\mathcal{B}}}
\newcommand{\cL}{{\mathcal{L}}}
\newcommand{\cS}{{\mathcal{S}}}
\newcommand{\pluseq}{\mathrel{+}=}

\clearpage
\section{Supplementary Material for Submission:
Towards General and Autonomous Learning of Core Skills - A Case Study in Locomotion}


\subsection{Method Details and Hyper Parameters}
\label{sec:experimental_details}

As defined in \cite{sacx1}, the problem of \RLD in a \MDPD is considered.
Let $\bs \in \mathbb{R}^S$ be the state of the agent in the MDP $\cM$, $\ba \in \mathbb{R}^A$ the continuous action vector and $p(\bs_{t+1} | \bs_t, \ba_t)$ the probability density of transitioning to state $\bs_{t+1}$ when executing action $\ba_t$ in $\bs_t$. 
All actions are assumed to be sampled from a policy distribution $\pi_\btheta(\ba | \bs)$, with parameters $\btheta$.
With these definitions in place, we can define the goal of Reinforcement Learning as maximizing the sum of discounted rewards $\bE_{\pi} \lbrack R(\tau_{0:\infty}) \rbrack = \bE_{\pi} \lbrack \sum_{t=0}^\infty \gamma^t r(\bs_t, \ba_t) \mid a_t \sim \pi(\cdot | \bs_t),\, \bs_{t+1} \sim p(\cdot | \bs_t, \ba_t),\, \bs_0 \sim p(\bs)  \rbrack$, where $p(\bs)$ denotes the state visitation distribution, and we use the short notation $\tau_{t:\infty} = \lbrace (\bs_t, \ba_t), \dots \rbrace$ to refer to the trajectory starting in state $t$.\\
The main idea of the multi-task RL setting in Scheduled Auxiliary Control (SAC-X) \cite{sacx1} is, that we have a main MDP $\cM$ and a set of auxiliary MDPs $\sA = \lbrace\cA_1, \dots, \cA_K\rbrace$.
These MDPs share the state, observation and action space as well as the transition dynamics, but have separate reward functions $r_{\cA_1}(\bs, \ba), \dots, r_{\cA_K}(\bs, \ba), r_{\cM}(\bs, \ba)$.
After executing an action -- and transitioning in the environment -- the agent now receives a scalar reward of all the auxiliary rewards and the main reward.\\
Given the set of reward functions we can define intention policies and their return as
$\pi_\btheta(\ba | \bs, \cT)$ and
\begin{equation}
\bE_{\pi_\btheta(\ba | \bs, \cT)} \Big\lbrack R_\cT(\tau_{t:\infty}) \Big\rbrack = \bE_{\pi_\btheta(\ba | \bs, \cT)} \Big\lbrack \sum_{t=0}^\infty \gamma^t r_\cT(\bs_t, \ba_t) \Big\rbrack,
\end{equation}
where $\cT \in \sT = \sA \cup \lbrace \cM \rbrace$, respectively.\\
Optimization of the policy is achieved by using an off-policy, model free RL approach, by trying to find an optimal multi-task value function $Q_\cT(\bs_t, \ba_t)$ for task $\cT$ as 
\begin{equation}
Q_\cT(\bs_t, \ba_t) = r_\cT(\bs_t, \ba_t) + \gamma  \bE_{\pi_\cT} \Big\lbrack R_\cT(\tau_{t+1:\infty}) \Big\rbrack,
\end{equation}
with ${\pi_\cT=\pi_\btheta(\ba | \bx, \cT)}$. 
Leading to the the (joint) policy improvement objective as finding $\arg \max_\btheta \cL(\btheta)$ where $\theta$ is the collection of all intention parameters and,
\begin{alignat}{3}
&\ \ \ &\cL(\btheta) &= \cL(\btheta; \cM) + \sum_{k=1}^{|\sA|} \cL(\btheta; \cA_k), \\
&\text{with}\ \ &\cL(\btheta; \cT) &= \sum_{\cB \in \sT} \mathop{\bE}_{p(s | \cB)} \Big\lbrack Q_\cT(\bs, \ba) \mid \ba \sim \pi_\btheta(\cdot | \bs, \cT)  \Big\rbrack.
\label{eq:obj_pol}
\end{alignat}
To optimize the objective a gradient based approach is used.
Using a parameterized predictor $\hat{Q}^\pi_\cT(\bs, \ba; \phi)$ (with parameters $\phi$) of state-action values; i.e. $\hat{Q}^\pi_\cT(\bs, \ba; \phi) \approx Q^\pi_\cT(\bs, \ba)$ and a replay buffer $B$ containing trajectories $\tau$ gathered from all policies, the policy parameters $\btheta$ can be updated by following the gradient
\begin{equation}
\nabla_\btheta \cL(\btheta) \approx \sum_{\substack{\cT \in \sT \\ \tau \sim B}} \mathop{\nabla_{\btheta} \bE}_{\substack{\pi_\btheta(\cdot | \bs_t, \cT) \\ \bs_t \in \tau}} \Big\lbrack \hat{Q}^{\pi}_\cT(\bs_t, \ba; \phi) - \alpha \log \pi_\btheta(\ba | \bs_t, \cT) \Big\rbrack,
\label{eq:policy_improve}
\end{equation}
where $\bE_{\pi_\btheta(\cdot | \bs_t, \cT)} \lbrack - \log \pi_\btheta(\ba | \bs_t, \cT) \rbrack$ corresponds to an additional (per time-step) entropy regularization term (with weighting parameter $\alpha$).\\
The second step in \cite{sacx1} is to find an optimal schedule during training that allows to learn the main task $\cM$ in a data-efficient way by executing the auxiliaries to collect appropriate data and help with exploration.
To achieve this, the scheduler divides an episode in a number of subsequent sequences and decides which intention is executed in a certain sequence. 
In \cite{sacx1} two schedulers are proposed, a pure uniform random scheduler, called SAC-U, and an optimizing scheduler SAC-Q.

To recap, we can apply the approach from \cite{sacx1} in three different ways:
\begin{itemize}
    \item In a single-task setting e.g. $\sT = \sA = \lbrace\cA_{\TASKWALKFWD}\rbrace$ , where the approach simply reduces to an off-policy RL experiment. This is used in the first set of experiments to show the properties of the skill rewards in section (\ref{sec:single_task_experiments} and \ref{sec:walking_blind}).
    \item In a multi-task setting with a set of locomotion skills, where we show that we can learn a set of auxiliaries in parallel, e.g. $\sT = \lbrace\cA_{\TASKWALKFWD}, \cA_{\TASKUPRIGHT}, \cA_{\TASKWALKBWD}\rbrace$
    but without using a main task and the random uniform scheduler (see section \ref{sec:sac_experiments_without_main}).
    \item Or in the full setting in section \ref{sec:sac_main_task}, where we have the complex and sparse task \TASKREACH and a set of auxiliaries that will help to learn it e.g. $\sT = \lbrace\cA_{\TASKWALKFWD}, \cA_{\TASKUPRIGHT}, \cA_{\TASKWALKBWD}\rbrace \cup \lbrace \cM_{\TASKREACH} \rbrace $, using the full SAC-Q scheduler setup in \cite{sacx1}.
\end{itemize}


We use the same hyper parameters for all experiments. Following \cite{sacx1} the stochastic policy consists of a layer of 256 hidden units with an ELU activation function, that is shared across all intentions.
After this first layer a layer norm is placed to normalize activations.
The layer norm output is fed to a second shared layer with 256 ELU units.
The output of this shared stack is routed to a head network for each of the intentions.
The heads are built from a layer of 100 ELU units followed by another layer of ELU units and a final tanh activation with twice the number of action dimension outputs, that determine the parameters for a normal distributed policy (whose variance we allow to vary between 0.3 and 1 by transforming the corresponding tanh output accordingly).
For the critic we use the same architecture, but with 400 units per layer in the shared part and a 300-1 head for each intention.
Training of both policy and Q-functions was performed via using a learning rate of $2 \cdot 10^{-4}$ (and default parameters otherwise), a discount factor of 0.99 and a replay buffer size of four million.
\\
For each of the simulation experiments the agent interacts with the simulated environment on episodes with a data rate that makes it comparable to experiments on a single real robot (single actor).
For all experiments we run two sequences of 400 steps with a step duration (of the simulated physics) of 25 milliseconds.
This gives us 20 seconds of simulated interaction in each of the episodes overall.
In each episode we measure the accumulated intention reward over the first sequence for the executed intention policy.
To measure the performance, we average the accumulated intention reward for the last 10 episodes for which that specific intention was active in the first sequence. In this way we measure the performance from the set of starting states.
For each task, we report the average number of episodes we need to have this performance measure exceed a threshold (or convergence, whatever happens first) over three independent seeds. To be able to compare a single reward definition over different robot platforms, we use the same task specific threshold for all platforms.
The threshold is chosen so that we see a minimal expected behaviour (average speed of 0.1 m/s for the walk tasks, 0.05 rad/s for the turn task, average height of 1 cm for the feet) without exceeding a roll or pitch angle of $\pm0.4$ radians.
We use the same procedure to report the episodes for the real robot experiments, but we run only one experiment (not several seeds) for each experiment in the real world.
It is also important to note, that if we report a certain number of episodes for the multi-task experiments, we report all episodes the agent interacted with environment to learn all the tasks from scratch (not per task).


\subsection{Reward Details}
\label{sec:reward_details}

We assume that all robots have access to an IMU which allows them to estimate the roll and pitch angles of the robot w.r.t. gravity.
In contrast to the roll and pitch angles, which can be reliably estimated from accelerometer and gyroscope data, the absolute yaw angle of the robot is typically estimated based on the earth's magnetic field, which especially indoors is often disturbed by other electromagnetic devices (including the robot's motors themselves) and hence unreliable. However this does not represent a problem since basic locomotion skill should be invariant w.r.t. to the yaw angle.\\
Using these measurement, this allows us to work in a virtual reference coordinate system $F_{H}$ that is simultaneously aligned with the robots forward direction and gravity.
Hence $F_{H}$ has the same origin as the torso coordinate frame $F_{T}$, has a x-y plane parallel to the worlds x-y plane, and no yaw component w.r.t. $F_{T}$.
This reference frame $F_{H}$ allows simple computation of different rewards that can be used over a broad range of different walker topologies.
Drawing on the forward kinematics of the walker, we represent each foot of the walker as a set of reference points (1 point for spherical feet, 8 corner points for plate feet).
Using the IMU, joint angles and forward kinematics, we can then compute the position of these reference points $j$ in the frame $F_{H}$ in each time step and for each foot $i$: $f^{H}_{ij}(t)$.\\
We reduce this to a single reference point for each foot $i$ by taking the reference point with the smallest z coordinate: $f^{H}_{i}(t) = f^{H}_{ij}(t)$ with $j = argmin_{j}(f^{H}_{ij}(t) \cdot (0, 0, 1))$.
We can also define a translational velocity of the feet reference points as $\delta f^{H}_{i}(t) = \frac{f^{H}_{i}(t) - f^{H}_{i}(t-dt)}{dt}$, where we neglect a small change in yaw between the consecutive coordinate frames.

To make use of these quantities we make the assumption that in each time step the robot is in contact with the ground and that the contact point is close to the lowest reference point.
Using this assumption we can make an estimate of the translational torso velocity relative to the world as $\delta T = -\delta f^{H}_{a}(t)$ with $a = argmin_{i}(f^{H}_{i}(t) \cdot (0, 0, 1))$.

\subsubsection{\TASKUPRIGHT}

We first define a reward function that encourages the robot to stay upright and not to fall or lean the torso in any direction.
Using our proprioceptive definitions and measurements, we first define a reward term to keep roll and pitch angle small.
Given roll angle $\phi(t)$ and pitch angle $\theta(t)$ we define this reward as:
\begin{equation}
  r_{up}(t) = 1 - c_{prec}(\sqrt{\phi(t)^2 + \theta(t)^2}, 0.0, 0.4)
\end{equation}
Given a general precision cost function:
\begin{equation}
  \begin{array}{c}
  c_{prec}(v, t, m) = \tanh{|(v-t) * w|}^{2} \\[\jot]
  w = \frac{atanh(\sqrt{0.95})}{m}
  \end{array}
\end{equation}
In addition we want to punish movements of the torso relative to the ground.
Assuming that we can estimate the torso velocity relative to the ground in the x and y axis of $F_{H}$ as $v_{xy}$ (taken directly from $\delta T$), we have:
\begin{equation}
  r_{still}(t) = -|v_{xy}(t)|
\end{equation}
As a last component of the reward, we want to prevent the torso from rotating. Assuming that we can measure the torso rotation rate directly from the gyroscope as $g_z(t)$, we can formulate this reward component as a negative thresholding of another reward r:
\begin{eqnarray}
  r_{rot}(t, r) = min( k*r , r)\\
  \text{with } k = 1.0 - c_{precise}(\hat{g}_Z(t), 0.0, 0.5)
\end{eqnarray}
Given these definitions, we can now define the \TASKUPRIGHT reward as:
\begin{equation}
  r_{\TASKUPRIGHT}(t) = r_{rot}(t, r_{still}(t) + r_{up}(t))
\end{equation}

\subsubsection{\TASKTURN}
For the turn task we expect the robot to rotate as fast as possible around the z axis of the torso while being upright.
Using the already given reward terms, we directly increase the gyroscope value $g_z(t)$ (instead of punishing, as we did in the \TASKUPRIGHT reward) while still keeping the torso levelled.
\begin{eqnarray}
  r_{\TASKTURN}(t, dir) = dir * g_z(t) + 0.1 * r_{up}(t) \\
\end{eqnarray}
In this investigation we consider turning left and right:
\begin{eqnarray}
  r_{\TASKTURNLEFT}(t) = r_{\TASKTURN}(t, 1.0) \\
  r_{\TASKTURNRIGHT}(t) = r_{\TASKTURN}(t, -1.0) \\
\end{eqnarray}

\subsubsection{\TASKLIFTFOOT}
For the task of lifting a certain foot $i$, $\TASKLIFTFOOT_i$, we define a reward, $r_{\TASKLIFTFOOT}(t, i)$, that tries to stand still while lifting a certain foot $i$ over a threshold of 5 cm. We use the definitions from before and add an $r_{lift}$ incentive:
\begin{equation}
  r_{\TASKLIFTFOOT}(t, i) = r_{rot}(t, r_{lift}(t, i) + 0.1 * r_{still} + 0.1 * r_{up}(t))
\end{equation}
with $r_{lift}$ being a bounded shaped reward of the height of the foot relative to the stand leg:
\begin{equation}
  r_{lift}(t, i) = min(1, h)
\end{equation}
\begin{eqnarray}
  h = (f^{H}_{i}(t) \cdot (0, 0, 1)) - (f^{H}_{a}(t) \cdot (0, 0, 1)) \\
  \text{ a being stand leg id}
\end{eqnarray}

\subsubsection{\TASKWALK}

Finally we define a reward for moving in a certain direction $\hat{v}_{xy}$ (with $\|\hat{v}_{xy}\| = 1$), relative to the x-y-plane of $F_{H}$.
To compute the full reward for robust locomotion only based on robo-centric measurements, we define a reward term for moving the torso in the desired direction: $r_{torso}(t, \hat{v}_{xy})$.
\begin{equation}
  r_{torso}(t, \hat{v}_{xy}) = r_{rot}(t, \hat{v}_{xy} \cdot v_{xy})
\end{equation}
As we saw in our experiments in simulation and in the real word, adding another incentive to move legs in the same direction helps to increase robustness and data efficiency.
We define a foot swing velocity $v_{swing}^{i}$ in the frame $F_H$ as:
\begin{equation}
  v_{swing}^{i} = \delta f^{H}_{i}(t) + \delta T
\end{equation}
If we neglect the z coordinate, we can now also define an incentive to move the feet forward:
\begin{equation}
  r_{feet}(t, \hat{v}_{xy}) = r_{rot}(t, \frac{1}{|i|}\sum{\hat{v}_{xy} \cdot v_{swing}^{i}} )
\end{equation}
This will cause a small incentive to move feet forward. For all leg in contact with the ground, this will have neither a positive or negative reward. For legs moving with the torso the rewards grows.

Finally the reward for walking is defined by:
\begin{equation}
  \label{reward:Walk}
  r_{\TASKWALK}(t, \hat{v}_{xy}) = r_{torso}(t, \hat{v}_{xy}) + 0.5 * r_{feet}(t, \hat{v}_{xy}) + 0.1 * r_{up}(t) \\
\end{equation}

In this investigation we use 4 instances of this reward:
\begin{eqnarray}
  r_{\TASKWALKFWD}(t) = r_{\TASKWALK}(t, (1,0)) \\
  r_{\TASKWALKBWD}(t) = r_{\TASKWALK}(t, (-1,0)) \\
  r_{\TASKWALKRIGHT}(t) = r_{\TASKWALK}(t, (0,-1)) \\
  r_{\TASKWALKLEFT}(t) = r_{\TASKWALK}(t, (0,1)) \\
\end{eqnarray}


\subsection{Action and Observation Details}
\label{sec:observations_details}
As stated in the main paper, we use the position control mode of the HEBI actuation modules.
For convenience, the agent action is constrained to the range $d \in [\delta_{min}, \delta_{max}]$ and transformed to actuator position command $\hat{p}$ by adding an initial position $\alpha$: $\hat{p} = \alpha + d$ for each of the actuators.

\begin{table}[h!]
\centering
 \begin{tabular}{||c c c c||} 
 \hline
 Action & Unit & dim & range\\ [0.5ex] 
 \hline\hline
 position set point & [rad] & 1 & [$\alpha+\delta_{min}$, $\alpha+\delta_{max}$]\\ [1ex]
 \hline
 \end{tabular}
 \caption{Action space for each actuator.}
 \label{table:act_actions_default}
\end{table}

In table \ref{table:act_obs_default} we summarize the observations that are used for each of the HEBI actuation modules.
The raw values are sent with 400 Hz over a ROS node running on the robot, while the filter state of the set-point smoothing window filter (of length $\nu=5$ steps) is stored in the agent.
The commanded action is computed by updating the filter state with the agent action and communicating the mean value over the last $\nu$ steps to the actuation module.

\begin{table}[h!]
 \centering
 \begin{tabular}{||c c c||} 
  \hline
  Observation & Unit & dim\\ [0.5ex] 
  \hline\hline
  position & [rad] & 1 \\
  velocity & [rad/s] & 1 \\
  elastic element deflection & [rad] & 1\\
  elastic element deflection velocity & [rad/s] & 1\\
  winding temperature & [$^\circ$C] & 1\\
  housing temperature & [$^\circ$C] & 1\\
  filter state & [rad] & $\nu=5$\\ [1ex]
  \hline
 \end{tabular}
 \caption{Observations for each actuator module.}
 \label{table:act_obs_default}
\end{table}

Each actuation module provides a filtered orientation estimate, as well as acceleration and gyroscope readings, based on it's own IMU.
We use a simple kinematics equation for each of the creatures to compute these values for the torso, based on the estimates of all modules directly attached to it.
While already the estimate from a single modules would be sufficient, we can use multiple modules to make it more robust.

For the observation vector of the robot, we use a history of $h=2$ time steps of the roll and pitch angle estimates to capture also the first derivative of these values in the observation.
These values are all independent of the yaw angle, that is also computed by all modules.
As the yaw angle has typically not a reliable absolute reference if we only take internal measurements of the robot, we ignore it for the agent observations as well as for the reward calculations.
To capture the rotational velocities of the torso, we have access to the gyroscope readings of the fused IMUs.
\begin{table}[h!]
 \centering
 \begin{tabular}{||c c c||} 
  \hline
  Observation & Unit & dim\\ [0.5ex] 
  \hline\hline
  torso roll angle estimate & [rad] & $h$ x 1\\
  torso pitch angle estimate & [rad] & $h$ x 1\\
  feet relative positions estimate & [m] & $h$ x $n$ x $|j|$ x 3\\
  torso gyro values & [rad/s] & $h$ x 3\\[1ex]
  \hline
 \end{tabular}
 \caption{Additional robot observations.}
 \label{table:torso_obs_default}
\end{table}
As we only use robo-centric measurements, we also provide the feet points in the reference frame: $f^{H}_{ij}(t)$ (assuming $n$ feet with $|j|$ reference points each).

Table \ref{table:creatures_default} summarizes the range of different creatures with their respective observation and action dimensions that we use in this work.
For the foot reference points, we use a single point in the center of the sphere-like foot and eight points on the corners of the plate-like foot.

\begin{table}[h!]
\centering
 \begin{tabular}{||c c c c||} 
 \hline
 Creature & description & act dim & obs dim\\ [0.5ex] 
 \hline\hline
 \DAISYVI & hexapod, 6 legs & 18 & 244 \\ 
 \DAISYIV & quadruped, 4 legs & 12 & 166 \\
 \DOG & quadruped, 4 legs & 12 & 166 \\
 \DAISYIII & tripod, 3 legs & 9 & 127 \\
 \FLORENCE & biped, 2 legs & 12 & 238 \\
 \FLORI & biped, 2 legs & 12 & 238 \\
 \FLORIARMS & biped, 2 legs & 16 & 282 \\ [1ex]
 \hline
 \end{tabular}
 \caption{Basic creatures with respective action and observation space ($h=2$).}
 \label{table:creatures_default}
\end{table}


\subsection{Ablation: Robustness of Reward Definitions}
\label{sec:ablation_robustness}

As described in section \ref{subsec:rewards_general} our general reward scheme makes some basic assumptions that may appear to be pretty strict.
For example, given the lack of contact detection, we assume that the vertically lowest foot is always in contact with the ground. Another assumption is that this contact point does not slip on the ground.
In essence we use these assumptions to give the reward some semantics, while we do not expect them to be fulfilled at each point in time.
The central idea is that we also expect our method to still create useful locomotion behaviours if these assumptions are not fulfilled in each time step.

\subsubsection{Uneven Terrain}
\label{sec:ablation_walking_blind}
Walking over rough or cluttered terrain is a challenging and important task for locomotion.
Especially using only internal sensors and without any additional sensors like cameras, LIDAR or sensorized feet (e.g. with contact sensors). We created an environment with pedestals that will have a random height drawn uniformly between 0 and $h_{max}$ in each episode (see Figure \ref{fig:daisy6_tiled}), where the assumption that the lowest foot is always in contact with the ground will inevitably be violated.

The creature starts in the middle of the arena and has to solve the task \TASKWALKFWD, which it can only do by crawling over the pedestals.
Using the same reward definitions compared to the flat terrain experiments, the bipeds \FLORENCE and \FLORI are able to handle height differences of about $h_{max}=1 cm$.
As one can expect having more legs allows more robust behaviours.
\DAISYIV can handle height differences of $h_{max}=4 cm$ and gets stuck afterwards (mostly with it's hind feet).
Having even more legs helps not only by allowing for simple statically stable gaits, it also help in terms of redundant sensor information.
Consequently, \DAISYVI shows an even better performance and can handle up to $h_{max}=20 cm$.
The learned gate shows an interesting pattern that looks like it would "feel" it's way in the blind.

\subsubsection{Hybrid Locomotion}
As an additional ablation to show the versatility of our approach, we changed the topology of creatures in our zoo to be even more dynamic.
As shown in Figure \ref{fig:florence_skates} we replaced the foot plates of the bipeds \FLORENCE and \FLORI with passive skates (similar to inline skates).
To allow for the very same rewards and observations as before, we put 4 foot reference points on each of the wheels outer diameter. As each foot has 2 wheels, we have a total of 8 reference points that now rotate with the passive wheels.
When we do this, we can run the very same setting as in the previous experiment and do the same computations. The only difference is that we have to measure the passive wheel velocities to compute the location of the reference points.
To have a better comparability we don't put these in the observations and only use the reference points as described above.
We can learn \TASKWALKFWD, \TASKWALKBWD and \TASKUPRIGHT with a comparable number of interactions for the bipeds in the single and multi-task setting, while the motion of the robot looks completely different.
It learns a very dynamic skating behaviour, to stop and turn, just using the simple rewards we used in all of the experiments.

\end{document}